\documentclass[review]{elsarticle}%

\makeatletter
\def\ps@pprintTitle{%
 \let\@oddhead\@empty
 \let\@evenhead\@empty
    \def\@oddfoot{\footnotesize\itshape
         {Preprint Under Review} \hfill\today}%
 \let\@evenfoot\@oddfoot}
\makeatother

\usepackage{amsmath}
\usepackage{amsfonts}
\usepackage{amssymb}
\usepackage{amsthm}
\usepackage{bm}
\usepackage{indentfirst}
\usepackage{graphicx}
\usepackage{subfigure}
\usepackage{array}
\usepackage{fullpage}

\DeclareMathAlphabet{\mathsfsl}{OT1}{cmss}{m}{sl}

\newcommand{\PreserveBackslash}[1]{\let\temp=\\#1\let\\=\temp}
\newcolumntype{C}[1]{>{\PreserveBackslash\centering}p{#1}}
\newcolumntype{R}[1]{>{\PreserveBackslash\raggedleft}p{#1}}
\newcolumntype{L}[1]{>{\PreserveBackslash\raggedright}p{#1}}

\numberwithin{equation}{section}

\theoremstyle{definition}

\usepackage{anyfontsize}
\usepackage{enumerate}
\usepackage{esint}
\usepackage{etoolbox}
\usepackage{isomath}
\usepackage{mathrsfs}
\usepackage{mathtools}
\usepackage{multicol}
\usepackage{pgfplots}
\usepackage{scalerel}
\usepackage{stackengine}
\usepackage{tabulary}
\usepackage{tikz}

\makeatletter
\newcommand*\bdot{\mathpalette\bdot@{.65}}
\newcommand*\bdot@[2]{\mathbin{\vcenter{\hbox{\scalebox{#2}{$\m@th#1\bullet$}}}}}
\makeatother

\makeatletter
\newcommand*\bddot{\mathpalette\bddot@{.65}}
\newcommand*\bddot@[2]{\mathbin{\vcenter{\hbox{\scalebox{#2}
    {$\m@th#1\smash{{}_{\bullet}^{\bullet}}$}}}}}
\makeatother

\newcommand{\circled}[2][]{%
  \tikz[baseline=(char.base)]{%
    \node[shape = circle, draw, inner sep = .5pt]
    (char) {\phantom{\ifblank{#1}{#2}{#1}}};%
    \node at (char.center) {\makebox[0pt][c]{#2}};}}
\robustify{\circled}

\stackMath
\newcommand\reallywidecheck[1]{%
\savestack{\tmpbox}{\stretchto{%
  \scaleto{%
    \scalerel*[\widthof{\ensuremath{#1}}]{\kern-.6pt\bigwedge\kern-.6pt}%
    {\rule[-\textheight/2]{1ex}{\textheight}}
  }{\textheight}%
}{0.5ex}}%
\stackon[1pt]{#1}{\scalebox{-1}{\tmpbox}}%
}

\usepackage[utf8]{inputenc}
\usepackage{bm}
\usepackage{mathrsfs}
\usepackage{amsmath}
\usepackage{amsfonts}
\usepackage{amsthm}
\usepackage[ruled]{algorithm} 
\usepackage[noend]{algpseudocode} 
\usepackage{color}
\usepackage{setspace}
\usepackage{float}
\usepackage{import}
\usepackage{url}
\usepackage{multicol}
\usepackage{multirow}
\usepackage{subfigure}
\biboptions{sort&compress}
\usepackage[top=1in,bottom=1in,left=1in,right=1in]{geometry}

\usepackage{tikz}

\newcommand{\real}{\mathbb{R}}

\newcommand{\mcN}{\mathcal{N}}

\usepackage{xcolor}

\def \xb{\bm{x}}

\def \xib{{\boldsymbol\xi}}

\usepackage{lineno}
\usepackage[hidelinks]{hyperref}

\begin{document}
\begin{frontmatter}

\title{Large language models, physics-based modeling, experimental measurements: the trinity of data-scarce learning of polymer properties}


\address[nl]{Global Engineering and
Materials, Inc., Princeton, NJ 08540, USA}
\address[yy]{Department of Mathematics, Lehigh University, Bethlehem, PA 18015, USA}
\address[bl]{Department of Mechanical Engineering, Virginia Tech, Blacksburg, VA 24060, USA}
\address[sn]{Department of Fire Protection Engineering, University of Maryland, College Park, MD 20742, USA}

\author[nl]{Ning Liu}
\author[yy]{Siavash Jafarzadeh}
\author[bl]{Brian Y. Lattimer}
\author[sn]{Shuna Ni}
\author[nl]{Jim Lua}
\author[yy]{Yue Yu\corref{cor1}}\ead{yuy214@lehigh.edu}

\cortext[cor1]{Corresponding author}

\begin{abstract}

Large language models (LLMs) bear promise as a fast and accurate material modeling paradigm for evaluation, analysis, and design. Their vast number of trainable parameters necessitates a wealth of data to achieve accuracy and mitigate overfitting. However, experimental measurements are often limited and costly to obtain in sufficient quantities for finetuning. To this end, we present a physics-based training pipeline that tackles the pathology of data scarcity. The core enabler is a physics-based modeling framework that generates a multitude of synthetic data to align the LLM to a physically consistent initial state before finetuning. Our framework features a two-phase training strategy: (1) utilizing the large-in-amount while less accurate synthetic data for supervised pretraining, and (2) finetuning the phase-1 model with limited experimental data. We empirically demonstrate that supervised pretraining is vital to obtaining accurate finetuned LLMs, via the lens of learning polymer flammability metrics where cone calorimeter data is sparse.

\end{abstract}


\end{frontmatter}


\section{Introduction}


Hundreds of millions of tons of polymer materials have been crafted for use in a colossal and ever-expanding array of applications with unceasing novel material demands, spanning automotive and ship components \cite{tran2018fire}, consumer packaging \cite{silvestre2011food}, fabrics \cite{fu2023sustainable}, solar cells \cite{li2012polymer}, etc. The quest for qualified polymer materials for specific applications hinges on the precise prediction of their properties, driving the need for swift and accurate assessment methods. In this pursuit, machine learning (ML) based models \cite{doan2020machine,kazemi2024facilitating,kim2021polymer,tao2021benchmarking} have emerged as a promising platform that empowers more rapid and precise polymer property prediction compared to state-of-the-art analytical and numerical methods (e.g., density functional theory \cite{mcmullen1990density,wu2006density}). In particular, a variety of supervised learning methods such as graph-based neural models \cite{park2022prediction,gurnani2023polymer,liu2023ino} have excelled in predicting polymer property tasks, largely attributed to their awareness and explicit encoding of the underlying chemical topology. Nevertheless, the vast chemical space, together with the generally restricted availability of polymer property labels, has made supervised learning cumbersome. This, in turn, has fostered the necessity for generalizable molecular representation learning and further sparked the exploitation in the use of attention-based large language models (LLMs) \cite{ross2022large,kuenneth2023polybert,xu2023transpolymer} for polymer property prediction, where one can simply convert the intricate chemical structure to a compact string representation termed Simplified Molecular-Input Line Entry System (SMILES) \cite{weininger1988smiles,goh2017smiles2vec,ozturk2018deepdta,paul2018chemixnet,shin2019self}. SMILES is essentially equivalent to a flattening of the chemical graph via a depth-first pre-order traversal of the spanning tree structure, and it thereby encodes the molecular graph in an implicit manner. Applying LLMs to polymer property prediction thus amounts to a chemical language learning task where the LLM endeavors to grasp the semantic and syntactic rules of the molecular structure leveraging various unsupervised learning schemes such as masked language modeling. The pretrained LLMs, akin to chemical linguists, can then be applied for downstream tasks via finetuning using labeled data \cite{wu2023molformer,maziarka2024relative,wang2024geometric,ross2024gp,belgodere2022cloud}. Pioneering work \cite{ross2022large,kuenneth2023polybert,xu2023transpolymer} has demonstrated the exceptional power of LLMs in predicting polymer properties and the correct geometrical interpretation of the spatial connectivity across substructures in a molecule.

On the other hand, although LLMs are powerful in aiding material assessment and novel design, the current application of LLMs resorts to directly finetuning a previously pretrained LLM with (likely limited) labeled data for downstream applications \cite{kuenneth2023polybert,xu2023transpolymer}. This task poses grand challenge, as the vast number of trainable parameters in LLMs (in the order of tens of millions \cite{wu2018moleculenet}) necessitates a substantial amount of experimental data to finetune the LLM in order to reach a desired level of accuracy. However, this is generally infeasible in many material design and assessment tasks, where conducting a large set of experiments is either prohibitively expensive or simply unfeasible. In this context, it calls for a highly effective training method to finetune LLMs with only a handful of experimental data.

\begin{figure}[!t]
\centering
\includegraphics[width=1.0\textwidth]{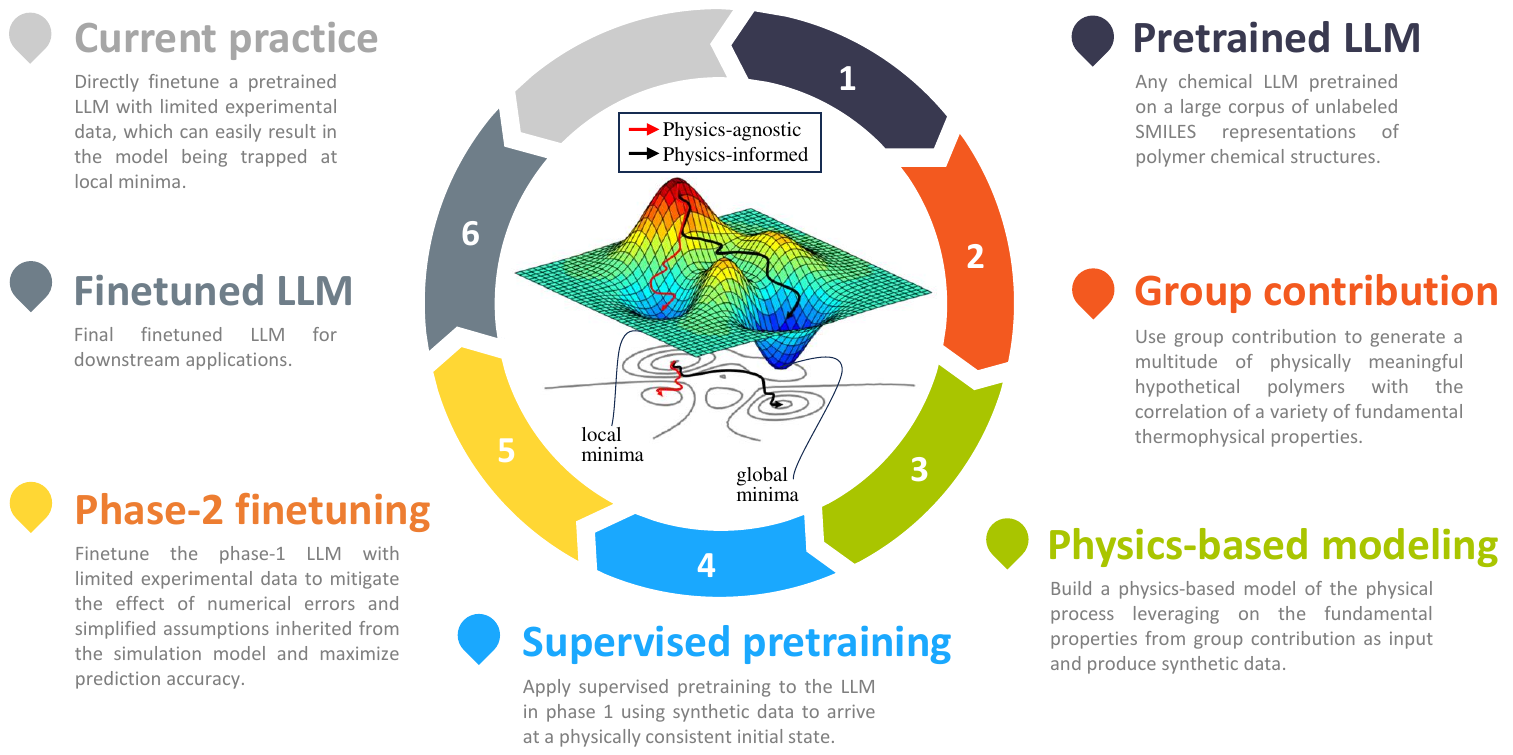}
\caption{An overview of the proposed trinity framework for data-scarce learning of LLMs.}
\label{fig:overview}
\end{figure}

A natural pathway to overcome this dilemma is to generate a large amount of synthetic data to pretrain the LLM, either in a supervised or unsupervised manner. Prior work \cite{kuenneth2023polybert} has explored decomposing existing polymers into chemical fragments \cite{degen2008art}, followed by randomly assembling these fragments to yield hypothetical polymers. While this method shows promise in pretraining the underlying LLM in an unsupervised manner, it is hard to judge whether the generated hypothetical polymers are physically meaningful, nor can these plausible polymers be correlated with any fundamental properties instrumental in constructing a physics-based simulation model or finetuning the final LLM decoder. In contrast, we take a different approach and investigate to incorporate physical knowledge into the synthetic polymer generation process. Specifically, we design a hypothetical polymer generation method grounded on physics-based group contribution (GC) \cite{lyon2009molecular}, based on which a variety of fundamental material properties such as the heat of combustion and heat capacity can be calculated. These computed fundamental properties can further be used to either finetune a decoder for direct downstream applications or serve as inputs to a numerical model of the physical asset to generate a large dataset of quantities of interest. An overview of the proposed trinity framework is illustrated in Figure~\ref{fig:overview}.

Besides the superiority in generating physically meaningful polymers, we show for the first time that, by simultaneously pretraining the encoder and decoder of the pretrained LLMs with physics-based synthetic data, 
the resulting LLMs inherit a better prior from GC, thereby achieving high predictability even in cases where experimental data is extremely limited. We then put forward a simple yet effective two-phase prediction-correction training strategy for finetuning pretrained LLMs towards polymer property prediction. In phase 1, we implicitly impose the physical knowledge from GC by training the encoder and decoder of pretrained LLMs using (possibly low-fidelity) physics-based synthetic data. As a result, the encoder and decoder parameters arrive at a physically consistent initial state. We then start from this initial state in phase 2 and further finetune the LLM with limited but most accurate experimental data. 
Analogous to pretraining the encoder of LLMs where one aims to train the encoder to learn the syntactical and grammatical rules of the underlying language \cite{kasneci2023chatgpt}, the rationale behind the first phase is to pretrain the encoder and decoder to understand the physical rules governing the underlying chemical formation of polymers, through a data-driven process fused with physics-based synthetic data. To distinguish our phase-1 pretraining from the unsupervised pretraining of LLM encoders, we term it \emph{supervised pretraining}. In phase 2, the limited experimental data is leveraged to mitigate the effect of numerical errors and simplified assumptions intrinsically inherited from the simulation model and maximize prediction accuracy.

Our main contributions are:
\begin{itemize}
\item{We propose a pathway that allows the generation of a large dataset of physically meaningful hypothetical polymers for LLM pretraining, anchored on the foundation of a physics-based GC with which a variety of fundamental properties can be correlated with the generated polymers.}
\item{
We put forth a physics-based modeling framework, leveraging on the fundamental properties obtained from GC as model input and efficiently producing a large amount of synthetic data.}
\item{We design a two-phase prediction-correction training strategy for finetuning LLMs in scientific problems in the absence of sufficient ground truth data, where the first phase applies supervised pretraining to the encoder and decoder using the generated (possibly less accurate) synthetic dataset, and the second phase employs the limited but most accurate experimental measurements for final finetuning/correction.}
\item{Interlocking the three pillars of LLMs, physics-based modeling and experimental measurements, we demonstrate the proposed trinity framework via the lens of learning polymer flammability metrics \cite{lyon2005polymer,kraft2016flammability,carpenter2005using,babrauskas1995development}, where cone calorimeter data is limited to dozens, and show that, by supervisedly pretraining the LLM with synthetic data, the model's prediction accuracy can be effectively improved by more than 50\%.}
\end{itemize}

To the authors' best knowledge, the present work represents the first attempt that explores the generation of physically meaningful synthetic polymers for supervised pretraining of LLMs. Together with the two-phase training strategy, we present not only an effective approach that makes the best use of limited labeled data, but also, more importantly, a comprehensive LLM learning framework applicable to a broad range of scientific problems.

\section{Results}\label{sec:results}

\textbf{Physically meaningful hypothetical polymer generation}

\begin{figure}[!t]
\centering
\includegraphics[width=1.0\textwidth]{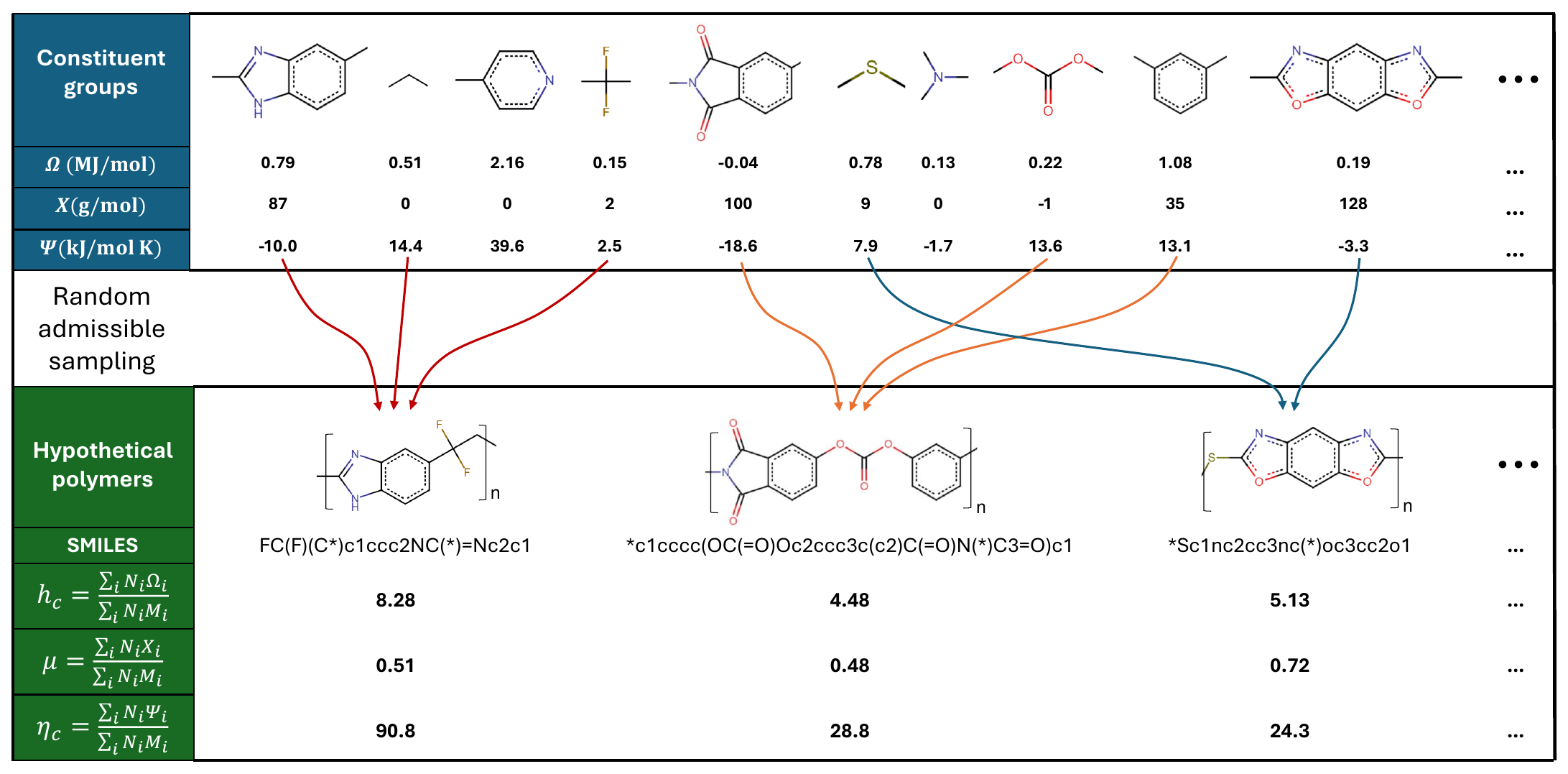}
\caption{Using group contribution to generate structurally admissible hypothetical polymers with physically meaningful material properties.}
\label{fig:group_contribution}
\end{figure}

We generate a set of structurally valid and physically meaningful hypothetical polymers based on GC \cite{lyon2009molecular,walters2003molar,hukkerikar2012group} (cf. the ``Methods'' section). The method provides a list of groups, their contributions to a variety of physical properties, and a relationship that estimates the compound's properties from the contributing groups. To estimate a property for a compound, one typically identifies the constituent sub-molecule groups from the list and uses their contributions and the relationship to obtain an estimation. In this work, we exploit a method to generate labeled hypothetical polymers. The idea is to use various admissible combinations between groups to construct new compounds and use \eqref{eq: GC} to label them with associated pyrolysis kinetic parameters. An admissible combination refers to a selection of groups that can form a structurally valid polymer compound (i.e., without open valences except at polymerization points). 
In the current study, we sample groups with two open valences and obtain unique combinations for up to three groups, which results in a total of 3237 hypothetical polymers. 
Figure~\ref{fig:group_contribution} demonstrates the formation of physically admissible polymers from constituent groups and the process of obtaining estimations for the associated pyrolysis properties.

\textbf{Physics-based modeling and synthetic data generation}

To generate adequate synthetic data for supervised pretraining of LLMs, a numerically accurate and computationally efficient model of the physical process needs to be constructed. In the instantiation to assessing the fire performance of polymers, we adopt the state-of-the-art fire simulation toolkit, Fire Dynamics Simulator (FDS) \cite{mcgrattan2013fire}, to emulate the cone calorimeter test and predict polymer flammability metrics such as the time to ignition $t_\text{ig}$ and the peak heat release rate per unit area pHRR. A brief introduction of the cone calorimeter test, FDS and our modeling strategy is provided in the ``Methods'' section. In this context, our goal is to first build a high-fidelity FDS model that correctly simulates the thermal decomposition process of polymers and the accompanied pyrolysis gas generation that further fuels the fire. With the correct physics being captured, we can then build a reduced-order model (ROM) for rapid data generation via either simplifying the high-fidelity model or ML.

\begin{figure}[!t]
\centering
\includegraphics[width=1.0\textwidth]{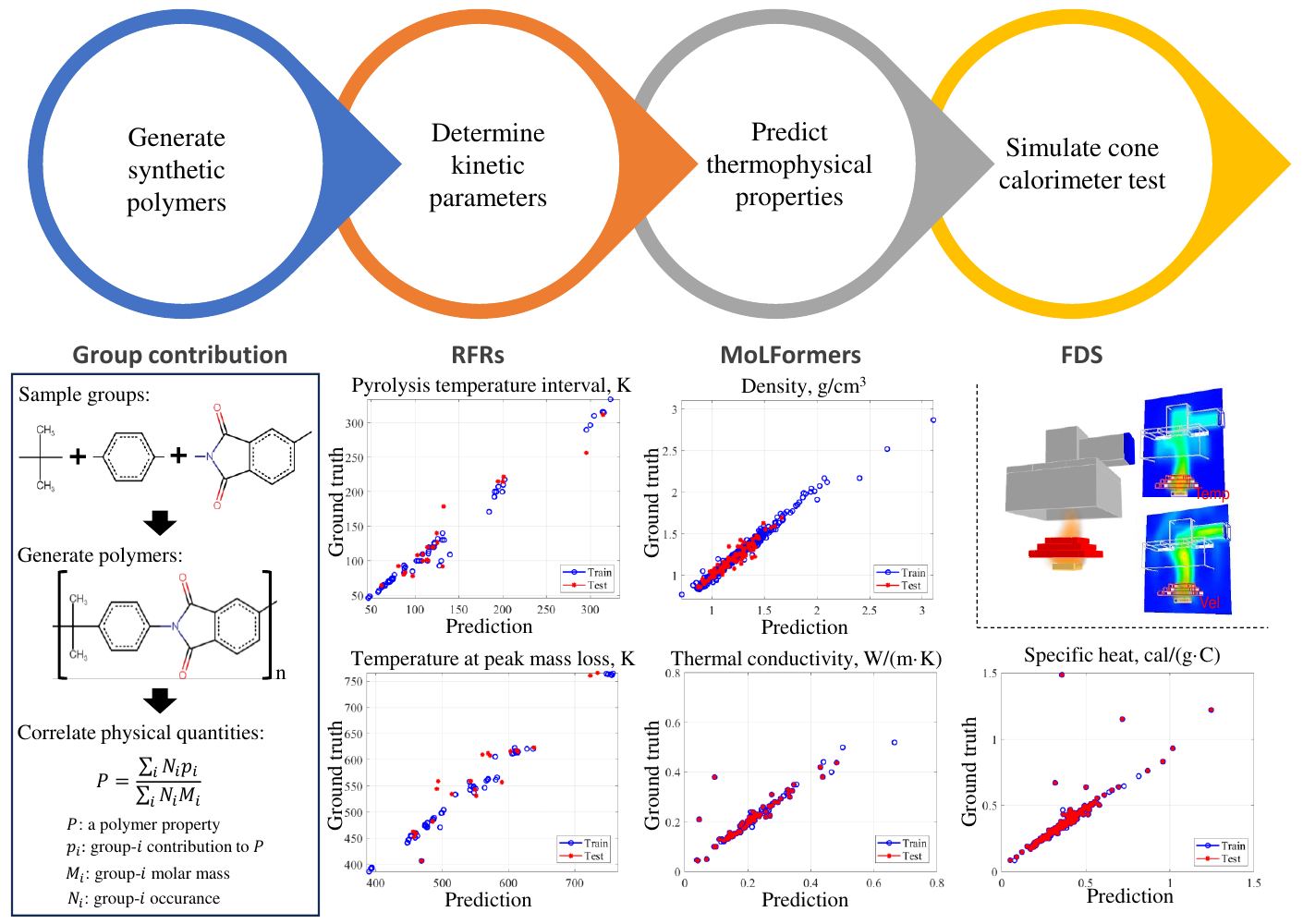}
\caption{Demonstration of the synthetic data generation workflow.}
\label{fig:workflow}
\end{figure}

 A demonstration of the workflow for physically meaningful synthetic data generation can be found in Figure~\ref{fig:workflow}. To simulate the cone calorimeter process in FDS, several material properties need to be defined, including thermophysical properties like density $\rho$, thermal conductivity $\kappa$ and specific heat $c_p$, and kinetic pyrolysis parameters like heat release capacity $\eta_c$, heat of combustion $h_c$, char fraction $\mu$, the temperature range of pyrolysis $dT$, and the temperature at peak mass loss rate $T_p$. Through GC for physically meaningful hypothetical polymer generation, the kinetic pyrolysis parameters of $\eta_c$, $h_c$ and $\mu$ are implicitly correlated. By further assuming single reaction, the temperature range of pyrolysis $dT$ is dependent on $\eta_c$ and $h_c$, however without an explicit algebraic connection \cite{lyon2009molecular}. Additionally, previous work \cite{lyon2009molecular} has shown that the temperature at peak mass loss rate $T_p$ is a function of $\eta_c$, $h_c$, and $dT$. We therefore employ two random forest regressors (RFRs) to learn these relationships, one from $(\eta_c, h_c)$ to $dT$ and the other from $(\eta_c, h_c, dT)$ to $T_p$. The two trained RFRs reach a $R^2$ accuracy of 0.99/0.92 and 0.99/0.88 for training/testing, respectively. Note that GC is a low-fidelity method and can only provide estimates. To improve the quality of the synthetic labeled data, we filter out nonphysical instances. In particular, we only keep the compounds with $\mu < 1$, $h_c<50$, $\eta_c>0$, and $1<dT<300$ and discard otherwise. After filtering, we come in possession of 3237 physically meaningful hypothetical polymers labeled with their kinetic pyrolysis parameters.

 On the other hand, since the thermophysical properties are relatively more available, we resort to the online polymer database, PolyInfo \cite{otsuka2011polyinfo}, to build three training datasets for density $\rho$, thermal conductivity $\kappa$ and specific heat $c_p$, covering various polymer types including homopolymers and copolymers. A summary of the employed datasets is listed in Table~\ref{tab:datasets}. With these three datasets, we take the SMILES representations of polymers as input and train three large chemical language models to predict $\rho$, $\kappa$ and $c_p$, respectively. Although the proposed framework is model-agnostic, in this work we employ MoLFormer \cite{wu2023molformer,ross2022large}, an open-source pretrained molecular transformer, for demonstration. We report the test errors in terms of the relative mean squared error (MSE), and the best MoLFormers obtain 3.27\%, 16.35\%, and 5.82\% for $\rho$, $\kappa$ and $c_p$, respectively. Note that only the density dataset is randomly split into 932/100 for training and testing, respectively. Due to the relatively small datasets for $\kappa$ and $c_p$ and the good coverage of polymer types, we use all the data for training and only use the homopolymer data for testing, so as to get an accurate model for interpolation without considerably compromising generalizability. Leveraging GC, two RFRs and three MoLFormers, all the needed FDS input parameters can be computed.

With the above capability to define the input parameters for FDS, we subsequently build a high-fidelity FDS model to simulate cone calorimeter tests at a constant heat flux exposure of 50 $kW/m^2$, taking into account the full solid and gas phases of the process. This model, albeit validated against available polymer cone calorimeter data and showed high accuracy \cite{hodges2023engineering}, is computationally slow in that it takes 1-4 hours to run one simulation. We then simplify the model by only considering the solid phase of the decomposition process with an additional simulation that includes a step increase in the heat flux of 25 $kW/m^2$ after ignition to account for the heat flux from the flames back down onto the polymer sample. This ROM produces comparable results to the high-fidelity model and can be completed in approximately a few seconds, thus enabling rapid generation of a large synthetic dataset for supervised pretraining.

\begin{table}[!h]
\caption{A summary of the employed datasets.
}
\label{tab:datasets}
\begin{center}
\centering
\begin{tabular}{llllr}
\hline
Property & Unit & Source & Data range & \# datapoints\\
\hline
Thermophysical &&&& \\
Density $\rho$ & $g/cm^3$ & PolyInfo & [0.770, 2.868] & 1032 \\
Thermal conductivity $\kappa$ & $W/(m\cdot K)$ & PolyInfo & [0.013, 23] & 110 \\
Specific heat $c_p$ & $cal/(g\cdot C)$ & PolyInfo & [0.085, 2.520] & 243 \\
\hline
Pyrolysis kinetic &&&& \\
Heat release capacity $\eta_c$ & $J/(g\cdot K)$ & Literature & [20.291, 1527.251] & 88 \\
Heat of combustion $h_c$ & $kJ/g$ & Literature & [3.823, 46.528] & 88 \\
Temperature range of pyrolysis $dT$ & $K$ & Literature & [46.093, 333.964] & 88 \\
Temperature at peak mass loss rate $T_p$ & $K$ & Literature & [386.285, 765.922] & 88 \\
\hline
Experimental cone calorimeter &&&& \\
Time to ignition $t_\text{ig}$ & $s$ & Literature & [1, 538] & 45 \\
Peak heat release rate pHRR & $kW/m^2$ & Literature & [19, 1761] & 45 \\
Smoke extinction area SEA & $m^2/kg$ & Literature & [33, 1300] & 38 \\
\hline
Synthetic cone calorimeter &&&& \\
Time to ignition $t_\text{ig}$ & $s$ & FDS & [20.2, 600] & 3237 \\
Peak heat release rate pHRR & $kW/m^2$ & FDS & [0.117, 1864.828] & 3237 \\
\hline
\end{tabular}
\end{center}
\end{table}

\textbf{Quantifying uncertainty of synthetic data}

\begin{figure}[!t]
\centering
\includegraphics[width=1.0\textwidth]{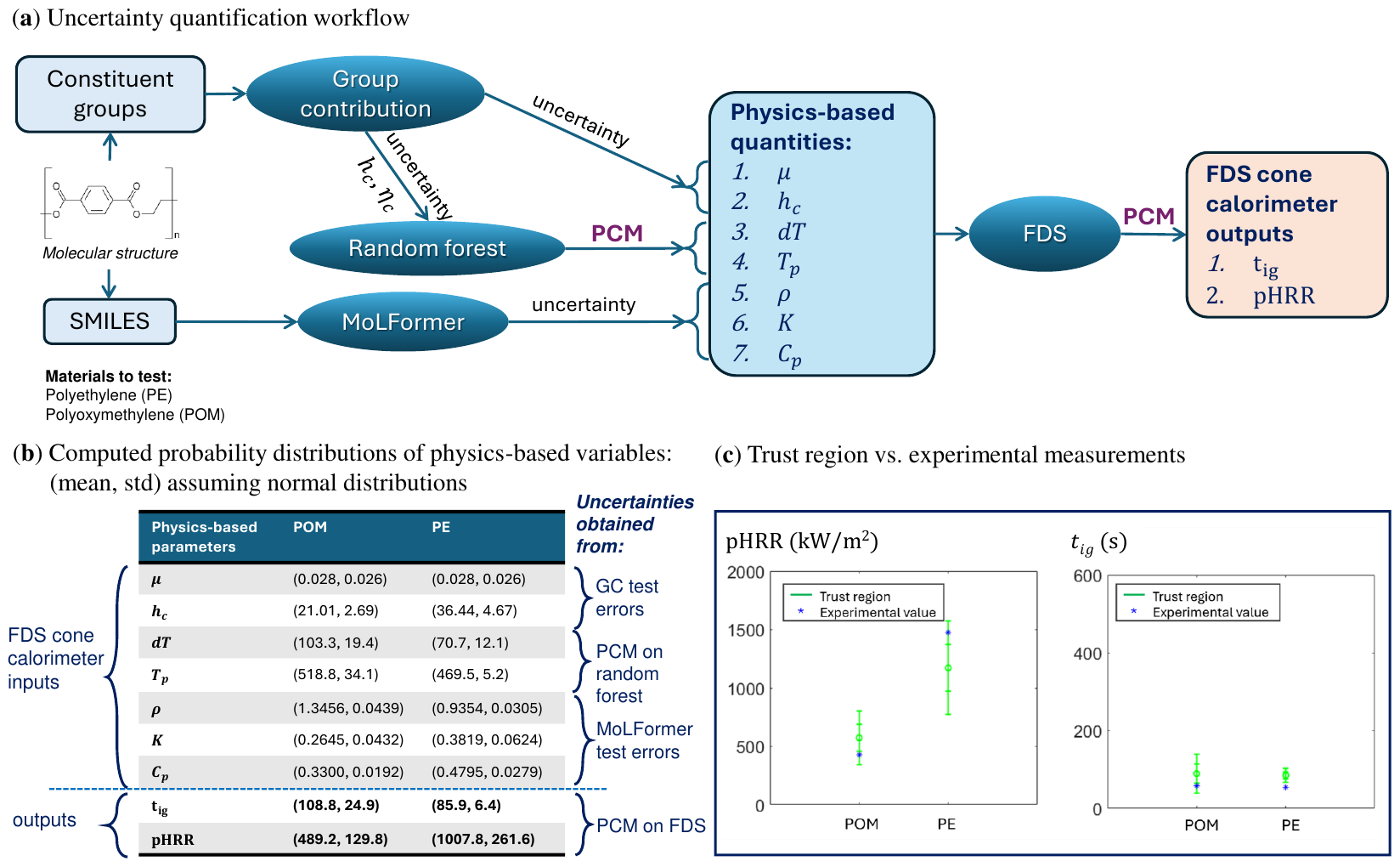}
\caption{Quantifying uncertainties on the generated synthetic data: (a) general workflow, (b) the computed probability distributions of physics-based variables, and (c) trust region vs. experimental measurement plots.}
\label{fig:uq}
\end{figure}

A key question is how reliable the generated synthetic data is. Both GC and RFR introduce unavoidable errors, which induce uncertainty in the FDS inputs. These uncertainties further propagate through the course of FDS simulation, resulting in posterior uncertainty in cone calorimeter predictions. To determine the trust regions for the synthetic $t_\text{ig}$ and pHRR of a particular polymer, we must quantify the uncertainty of FDS simulation from its inputs. In this regard, we use an additive independent and identically distributed (i.i.d.) noise to model the errors of FDS inputs. Specifically, we follow the practice in scientific ML \cite{moya2024conformalized,zou2023uncertainty,ghosh2023discovery} and assume that the distributions of all inputs follow Gaussian distributions, $\xi\sim\mcN(E_{\xi},\sigma^2_{\xi})$. The only exception is char fraction $\mu$, where a log-normal distribution is assumed: $\ln(\mu)\sim \mcN(E_{\ln(\mu)},\sigma^2_{\ln(\mu)})$, to guarantee the physical consistency of nonnegativity. For the kinetic parameters $\mu$, $h_c$, and $\eta_c$ obtained from GC, we approximate their means and standard deviations as the test results and the averaged test errors of GC. To estimate the distributions of $dT$ and $T_p$, we employ the probabilistic collocation method (PCM) \cite{xiu2005high,nobile2008anisotropic,ma2009adaptive,zhang2012error,lin2009efficient} on RFR models to determine the means and standard deviations. For the thermophysical inputs $\rho$, $K$, and $C_p$ generated via MoLFormers, we use MoLFormer's predictions and their averaged test errors to estimate their distributions. 
With the standard deviations for all seven FDS inputs obtained, we again apply PCM to the FDS model and find the distributions for $t_\text{ig}$ and pHRR. To verify this process, we consider two polymers from the synthetic dataset that correspond to known polymers with available experimental measurements, namely Polyethylene (PE) and Polyoxymethylene (POM). Figure~\ref{fig:uq}b lists the means and standard deviations for the FDS inputs and outputs. 
Figure~\ref{fig:uq}c displays the trust regions for $t_\text{ig}$ and pHRR predicted by the data generation method versus experimental measurements for PE and POM. The green segments span four standard deviations corresponding to 95$\%$ of the possible values. The inclusion of the measured values within or near the trust regions demonstrates the reliability and trust-worthiness of the synthetic data generation approach.

\textbf{Phase-1 supervised pretraining}

\begin{figure}[!t]
\centering
\includegraphics[width=1.0\textwidth]{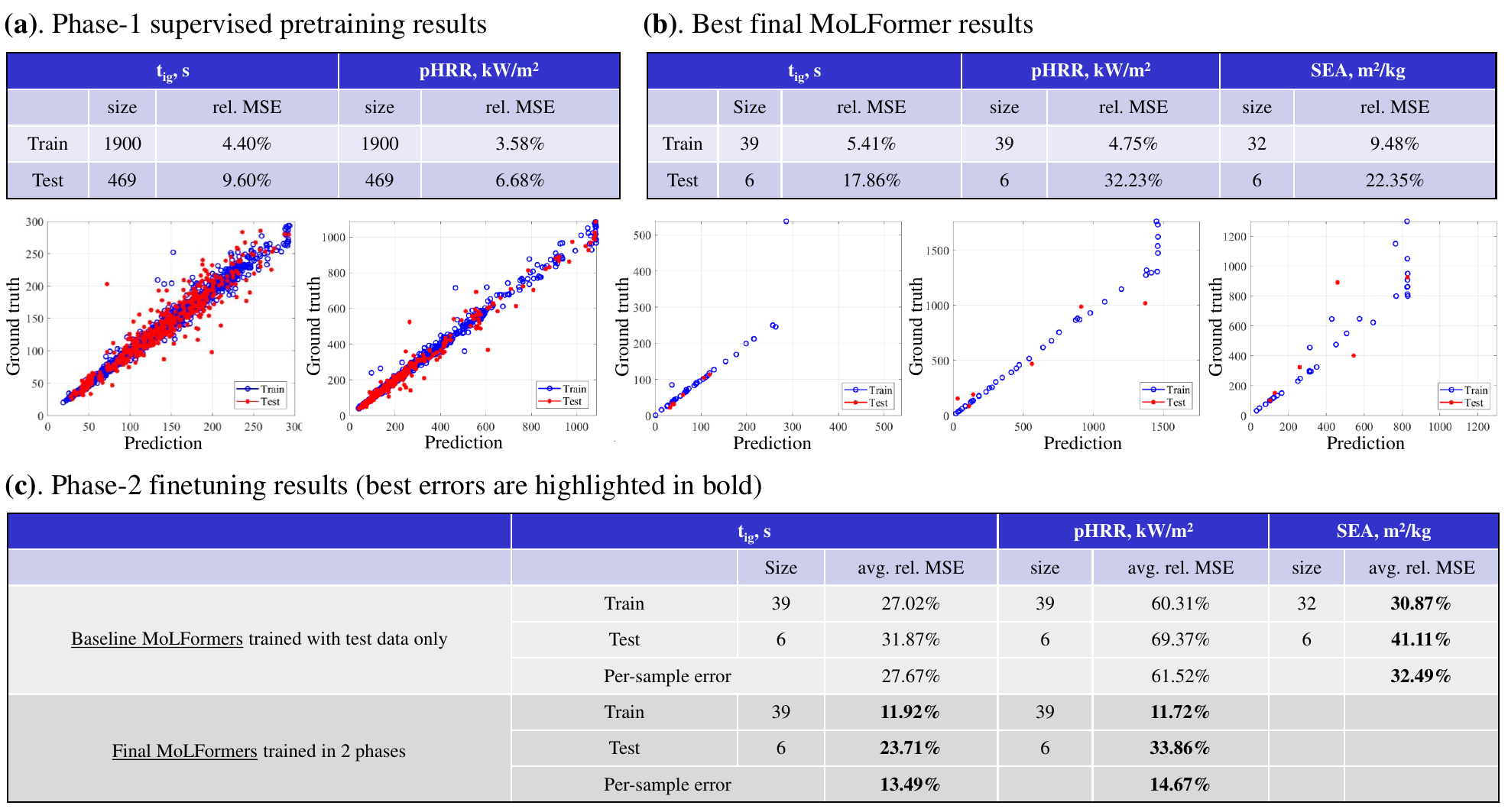}
\caption{MoLFormer training results for cone calorimeter metrics prediction.}
\label{fig:molformer_results}
\end{figure}

A significant advantage of the proposed trinity framework is the supervised pretraining of LLMs that aims to align the model to a physically consistent initial state prior to finetuning. In the circumstance of cone calorimeter test, we pretrain the LLM with labeled synthetic data in terms of time to ignition $t_\text{ig}$ and peak heat release rate per unit area pHRR. In particular, we first categorize the synthetic data into two classes, one with ignitable polymers (indicated by $t_\text{ig} \leq 600 s$) and the other with un-ignitable polymers. We slightly modify the MoLFormer architecture by adding a sigmoid function to the decoder output together with the binary cross-entropy loss as the loss function to train a classifier that identifies if a given polymer is ignitable. We then use the ignitable synthetic polymers for supervised pretraining. This pre-processing step is critical because un-ignitable polymers are characterized by a constant $t_\text{ig}$ of 600 s and very small pHRRs, which 
could otherwise introduce unbalanced outputs and deteriorate prediction accuracy. By implementing a pre-classification step, we address this data distribution challenge by focusing on the ignitable polymers of interest, thereby enhancing the model's predictability. Next, we randomly split the 2369 ignitable polymers into 1900 for training and 469 for testing. The supervised pretraining results can be found in Figure~\ref{fig:molformer_results}a, where the best trained MoLFormers achieve training and test errors of 4.40\%/9.60\% and 3.58\%/6.68\% for $t_\text{ig}$ and pHRR, respectively. With these superb training results, the two MoLFormers are anticipated to inherit good physical priors from the physics-based synthetic data and qualify as initial models for phase-2 finetuning.

\textbf{Phase-2 finetuning}

In phase 2, we finetune the obtained phase-1 models with good physics-based priors. Due to the limited size of the experimental cone calorimeter datasets (cf. Table~\ref{tab:datasets}), we randomly split each dataset 5 times and train 5 MoLFormers for both $t_\text{ig}$ and pHRR to mitigate the effect of bias in splitting data. We then report the averaged relative MSE in Figure~\ref{fig:molformer_results}c, together with the results from the best MoLFormers (out of the 5 trained) in Figure~\ref{fig:molformer_results}b. In general, we observe a good agreement between the ground truth data and the prediction, as indicated by their alignment along the $y=x$ line. We note that during training, the loss function is defined based on relative MSE. As such, the denominator becomes smaller for polymers with smaller pHRR and SEA values, which is equivalent to placing a larger weight on these polymers. Consequently, the model is anticipated to provide better predictions for polymers with lower flammability. This loss function is chosen because optimal polymer structures with lowest flammability are often more favorable \cite{bourbigot2006polymer}. 
The effect of the relative MSE loss can also be observed in our results shown in Figure~\ref{fig:molformer_results}, where the model generally presents larger errors in the higher pHRR and SEA regimes, underscoring the importance of an application-oriented loss function.
We also point out that with a proper selection of suitable loss functions or simply the augmentation of more data of large pHRRs and SEAs, the trained MoLFormers can be further finetuned to reach higher accuracy. This is, however, out of the scope of the current study.

\textbf{Enhanced predictability of the learned models}

To demonstrate the enhanced predictability of the final MoLFormers utilizing the 2-phase training approach, we compare their accuracy with the baseline MoLFormers trained with test data only in Figure~\ref{fig:molformer_results}c. Note that, due to the inability of FDS to generate synthetic SEA data, we do not compare the performance in terms of SEA. However, synthetic SEA data can be generated using other physics-based methods such as the smoke calculation method in \cite{tewarson1988smoke}, which is left as a future research direction. With this caveat in mind, by comparing qualitatively across the prediction plots in Figure~\ref{fig:molformer_results}b, a much wider dispersion from the $y=x$ line together with an early-stage deviation is observed in the SEA plot, emphasizing the indispensable need of the phase-1 supervised pretraining. By comparing the performance between the baseline single-phase MoLFormers (i.e., trained with test data only) and our final physics-guided MoLFormers in Figure~\ref{fig:molformer_results}c, the final MoLFormers manifest significantly higher accuracy in the prediction of both $t_\text{ig}$ and pHRR, where the test errors are improved by 25.6\% and 51.2\%, respectively. Note that, since the best models are chosen based on the best test errors, the training errors in the best models are also effectively reduced. By comparing the per-sample errors between the baseline and final MoLFormers, a remarkable improvement of 51.2\% and 76.2\% is observed in the prediction of $t_\text{ig}$ and pHRR, respectively. These results highlight the effectiveness of the proposed 2-phase training framework in imposing physics-based priors and tackling the pathology of data scarcity.

\section{Discussion}\label{sec:discussion}

We present a physics-guided ML pipeline tailored for finetuning LLMs with limited data availability, in which a 2-phase training strategy is designed to blend physics knowledge with pretrained LLMs and finite experimental measurements. Our training strategy begins with a supervised pretraining phase (phase 1), wherein physics-based synthetic data is leveraged to lead the LLM to a physically consistent initial state that establishes the prior physical knowledge for subsequent finetuning. In phase 2, the obtained phase-1 model, having acquired a condensed representation of polymers and a deep physical understanding of the downstream problem, undergoes finetuning that makes the best use of limited experimental data, thereby achieving the final LLM with enhanced precision.

The realization of the proposed 2-phase training framework relies on a suite of physics-based numerical models for efficient synthetic data generation. On one hand, group contribution plays a critical role, not only in generating a vast array of hypothetical polymers for LLM pretraining, but also in correlating each generated polymer with a diverse range of fundamental physical properties that materialize the hypothetical polymer with actual physical significance. On the other hand, the proposed physics-based modeling framework of the physical process enables the production of a multitude of synthetic data for supervised pretraining. Another outstanding feature of the 2-phase training strategy lies in its relaxation of the stringent accuracy requirement on synthetic data, as the phase-2 finetuning aims to utilize the most accurate experimental data to correct the phase-1 prediction and mitigate the impact of numerical errors and simplified assumptions intrinsically inherited from the numerical simulation model, thus maximizing prediction accuracy.

Interlocking the three pillars of scientific machine learning---LLMs, physics-based modeling and experimental measurements---we demonstrate the proposed trinity framework via the lens of learning polymer flammability metrics, where cone calorimeter data is limited (to only a few dozen measurements). We show that, through supervised pretraining of the LLM with synthetic data, the model's per-sample accuracy can be effectively enhanced by at least 50\%. This extraordinary capability in making accurate predictions while relaxing the requirement of an excessive amount of expensive experimental data enables the exploration of the gigantic polymer design universe and the discovery of optimal polymer structures for downstream applications in a precise and efficient manner.

\section{Methods}\label{sec:methods}

\textbf{SMILES canonicalization}

SMILES (Simplified Molecular Input Line Entry Systems) is developed in the 1980s \cite{weininger1988smiles} and can be viewed as a chemical language for describing molecular structures. While there can be more than one SMILES expressions for a given structure, canonical SMILES are unique. Canonicalization is an algorithm that maps any valid SMILES for a given structure to the unique (canonical) version for that molecule \cite{weininger1989smiles}.

Since the chemical structures of polymers can be described by their repeating units, researchers have adopted SMILES notations to represent polymers (e.g., \cite{BigSMILES} and \cite{PSMILES}). In this work, we employ a similar notation as in \cite{PSMILES}, where the polymerization point is denoted by an asterisk (i.e., ``*''). For copolymers, the corresponding SMILES consists of SMILES of the constituent monomers, connected by dots (i.e., ``.'').

\textbf{Group contribution}

The group contribution (GC) method is a low-fidelity methodology for estimating various properties of chemical compounds by dividing them into smaller constituents (i.e., sub-molecular structures), sometimes referred to as molar groups. A compound's property is then assumed to be computable according to the \emph{contribution} of the constituent molar groups \cite{joback1987estimation}. For any particular property, and given a dataset containing this property for a number of compounds, one needs to gather a finite set of molar groups that describe all the given compounds and any other compounds for which one aims to obtain the property. Then, a weighted linear equation is assumed that relates the property to the contribution of molar groups, and a regression problem is solved to obtain the group contributions. Once the contributions are found, they can be used to estimate the properties for compounds outside of the original dataset. 

The selection of molar groups and the equation that relates properties to the group contributions are not unique and vary across applications. In this work, we use the GC method by Lyon \cite{lyon2009molecular} that provides estimates for heat release capacity $\eta_c$, specific heat of combustion $h_c$, and char fraction $\mu$ for polymer compounds based on a set of 38 molar groups, using:
\begin{equation}
\label{eq: GC}
\eta_c = \sum_{i=1}^{n_g} \frac{N_i\Psi_i}{N_iM_i}\text{, }\quad
h_c = \sum_{i=1}^{n_g} \frac{N_i\Omega_i}{N_iM_i}\text{, }\quad
\mu = \sum_{i=1}^{n_g} \frac{N_iX_i}{N_iM_i}\text{, }
\end{equation}
where $n_g$ is the number of distinct groups in the compound, $N_i$ is the number of occurrence of group $i$ in the compound, $M_i$ is the molar mass of the group $i$, and $\Psi, \Omega_i, X_i$ are contributions of group $i$ to the $\eta_c, h_c, \mu$ properties, respectively.

\textbf{Numerical modeling of the physical process}

Two essential ingredients of a synthetic data generator are the numerical accuracy and computational efficiency of the simulation model. The model needs to be numerically validated against experimental measurements to produce trustworthy data. This is often achieved via constructing a high-fidelity model that simulates the complete process of the physical asset. Concurrently, the model should also be computationally efficient to be able to generate a large amount of data in a relatively short amount of time. This is often accomplished via reduced-order modeling of the high-fidelity process. 

We instantiate the modeling framework of the physical process in the form of cone calorimeter experiments \cite{babrauskas2016cone} that are used to study fire behaviors in small samples of condensed-phase materials. In this case, Fire Dynamics Simulator (FDS) \cite{mcgrattan2013fire} is adopted as the computational framework. FDS is a computational fluid dynamics (CFD) model based on large-eddy simulation developed by the National Institute of Standards and Technology, which predicts the thermo-chemo-physical response of the surrounding solid-phase materials and their potential contribution to the fire through the generation of pyrolysis gases that create fuel for the fire. In this context, we construct a high-fidelity FDS model to predict the fire behavior of polymers exposed to a constant heat flux from the cone heater. A high-fidelity simulation model is constructed that performs a single simulation to predict the temperature rise and the decomposition of polymers over time in the solid phase as well as the combustion of pyrolysis gases in the gas phase. This model, albeit being numerically accurate against experimental data, is computationally slow in that it takes 1-4 hours to run one simulation. To alleviate the computational burden, we proceed to build a reduced-order model (ROM) by only considering the solid-phase mode with an additional simulation that includes a step increase in the heat flux of 25 $kW/m^2$ after ignition to account for the heat flux from the flames back down onto the polymer sample. This ROM produces comparable results to the high-fidelity model and can be completed in approximately one second, thus enabling rapid generation of a large synthetic dataset for supervised pretraining.

\textbf{Probabilistic collocation for uncertainty quantification}

The Probabilistic Collocation Method (PCM) \cite{xiu2002wiener,xiu2005high,nobile2008anisotropic,ma2009adaptive,zhang2012error,lin2009efficient,fan2022asymptotically,fan2022meshfree} is a popular numerical method for stochastic models. It retains the ease of implementation found in Monte Carlo methods, as only solutions at the sample points are needed. This makes PCM particularly suitable in cases where the deterministic model operates as a black box or pre-compiled/pre-trained software. Additionally, when combined with sparse grids such as the Smolyak formulation \cite{smolyak1963quadrature}, PCM reduces the number of sample points needed to achieve a given numerical accuracy, especially for problems with relatively small ($\leq 50$) random dimensions \cite{lin2009efficient} and sufficient solution smoothness in the parameter space \cite{fan2022asymptotically}. These attributes make PCM an ideal candidate for our application.

We elaborate on PCM for a given stochastic model, $f(\xb,\xib)$, where $\xb\in\real^{d_x}$ denotes physical parameters and $\xib\in\Omega_\xi$ represents random variables of the random field. Here, $\Omega_\xi$ is the random space. It is often assumed that all elements of $\xib$ are i.i.d. random variables with a probability density $\rho:\Omega_\xi\rightarrow \real^+$. The goal of PCM is to construct an approximate solution manifold of $f(\xb,\xib)$ based on Lagrange interpolation in the random space. Specifically, let $\Theta_N=\{\xib_k\}_{k=1}^K\subset \Omega_\xi$ be a set of prescribed nodes such that the Lagrange interpolation can be performed in the random space $\Omega_\xi$, where $N$ is the dimension of the parametric space. Then, the function $f(\xb,\cdot):\Omega_\xi\rightarrow \real$ can be approximated using the Lagrange interpolation polynomial:
$$\mathcal{J}[f](\xb,\xib)=\sum_{k=1}^K f(\xb,\xib_k)J_k(\xib)\text{ ,} $$
where $J_k(\xib)$ is the Lagrange polynomial satisfying $J_k(\xib_j)=\delta_{kj}$. Then, the statistical moments of the stochastic model $f$ can be evaluated as follows:
\begin{equation}\label{eqn:pcm}
\mathbb{E}[f](\xb)\approx \int_\Gamma \sum_{k=1}^K f(\xb,\xib_k)J_k(\xib)\rho(\xib)d\xib\text{ , }\;\sigma[f](\xb)\approx \sqrt{\int_\Gamma \left[\sum_{k=1}^K f(\xb,\xib_k)J_k(\xib)\right]^2\rho(\xib)d\xib-[\mathbb{E}[f](\xb)]^2}\text{ ,}
\end{equation}
and so on. In our application, we take $f$ as the RFR model (when estimating the uncertainty of $dT$ and $T_p$) or the FDS model (when estimating the uncertainty in cone calorimeter tests). We evaluate the RFR and FDS models at each collocation point, $\xib_k$, producing a set of system responses corresponding to different input values. Using these responses, we can then estimate the statistical moments of the output distribution, such as the mean, variance, and higher-order moments, following \eqref{eqn:pcm}.

\textbf{Large chemical language models for polymer property prediction}

Without loss of generality, we employ the pretrained large chemical language model, MolLFormer \cite{ross2022large}, as our LLM backbone. MolLFormer is an efficient transformer encoder model with linear attention mechanism and rotary positional embeddings. Although the proposed trinity framework is applicable to any large chemical language model (i.e., it is LLM-agnostic), we choose MoLFormer for demonstration because it is pre-trained on a relatively large database of 1.1 billion unlabeled SMILES representations of compounds from the PubChem and ZINC datasets. This pretraining database covers a diverse range of real-world compounds and positions MoLFormer as an arguably more general-purpose chemical linguist. It has been shown that MolLFormer can capture an adequate amount of structural and chemical information for accurate predictions of a diverse array of downstream applications \cite{ross2022large}.

\textbf{Two-phase training strategy}

The two-phase prediction-correction training strategy consists of a supervised pretraining phase with a myriad of synthetic data, and a finetuning phase using limited experimental data. In the first phase, we aim to instil in the encoder and decoder the physical knowledge governing the underlying problem, via a data-driven process using a multitude of synthetic data. The rationale behind this phase closely resembles the unsupervised pretraining of transformer encoders in most LLMs that aims to pretrain the encoder to understand the syntactical and grammatical rules of the chemical language. Another significant advantage of the proposed strategy is that it relaxes the stringent requirement on the accuracy of the synthetic dataset. In fact, the phase-2 finetuning leverages the most accurate experimental measurements, treated as ground truth data, to correct the phase-1 predictions and mitigate the effects of numerical errors and simplified assumptions intrinsically inherited from the numerical simulation model. In this way, our two-phase training strategy enhances prediction accuracy by seamlessly blending prior physical knowledge with experimental measurements.

\section*{Acknowledgments}

This work is supported by the Office of Naval Research (ONR) under Grant No. N68335-24-C-0123. This support is gratefully acknowledged.



\end{document}